\documentclass[journal]{IEEEtran}
\ifCLASSINFOpdf
\else
\fi

\usepackage{graphicx}
\usepackage{comment}
\usepackage{amsmath,amssymb} 
\usepackage{color}
\usepackage{colortbl}
\usepackage{hyperref}

\newcommand{\tabincell}[2]{\begin{tabular}{@{}#1@{}}#2\end{tabular}}

\definecolor{mg1}{gray}{.5}
\definecolor{mg2}{gray}{.8}

\newcommand{\textfig}[1]{\small{#1}}

\newcommand{\para}[1]{}
\newcommand{\imgfile}[1]{./figs/#1.png}
\newcommand{\cjz}[1]{#1}

\newcommand{\xyh}[1]{#1}

\newcommand{\datasetname}{\textit{InstanceBuilding}}

\usepackage{multirow} 
\usepackage{array}
\begin{document}
%
\title{3D Instance Segmentation of MVS Buildings}
%
%
%

\author{Jiazhou~Chen,
        Yanghui~Xu,
        Shufang~Lu,
        Ronghua~Liang*,
        and~Liangliang~Nan
\thanks{J. Chen, Y. Xu, S. Lu and R. Liang are with the School of Computer Science and Technology, Zhejiang University of Technology, China}
\thanks{L. Nan is with the Deft University of Technology, The Netherlands}
\thanks{* Corresponding author}

}

%
%

\markboth{Journal of \LaTeX\ Class Files,~Vol.~59, No.~8, August~2021}%
{Chen \MakeLowercase{\textit{et al.}}: Bare Demo of IEEEtran.cls for IEEE Journals}
%



\maketitle

\begin{abstract}
We present a novel 3D instance segmentation framework for Multi-View Stereo (MVS) buildings in urban scenes. Unlike existing works focusing on semantic segmentation of urban scenes, the emphasis of this work lies in detecting and segmenting 3D building instances even if they are attached and embedded in a large and imprecise 3D surface model. Multi-view RGB images are first enhanced to RGBH images by adding a heightmap and are segmented to obtain all roof instances using a fine-tuned 2D instance segmentation neural network. Instance masks from different multi-view images are then clustered into global masks. Our mask clustering accounts for spatial occlusion and overlapping, which can eliminate segmentation ambiguities among multi-view images. Based on these global masks, 3D roof instances are segmented out by mask back-projections and extended to the entire building instances through a Markov random field optimization. \cjz{A new dataset that contains instance-level annotation for both 3D urban scenes (roofs and buildings) and drone images (roofs) is provided. To the best of our knowledge, it is the first outdoor dataset dedicated for 3D instance segmentation with much more annotations of attached 3D buildings than existing datasets\footnote{The datasets are available at \href{https://californiachen.github.io/datasets/InstanceBuilding}{https://californiachen.github.io/datasets/Instance- Building}}. Quantitative evaluations and ablation studies have shown the effectiveness of all major steps and the advantages of our multi-view framework over the orthophoto-based method.}
\end{abstract}

\begin{IEEEkeywords}
Instance segmentation, dataset, 3D urban scene, multi-view clustering.
\end{IEEEkeywords}

%
\IEEEpeerreviewmaketitle

\section{Introduction}
In recent decades, the Multi-View Stereo (MVS) technique has been widely used in the GIS domain. Multi-view images are captured by Unmanned Aerial Vehicle (UAV) and used to automatically reconstruct dense 3D mesh models of large urban scenes~\cite{vu2012high,yao2018mvsnet}. The reconstructed 3D mesh models provide a visually pleasing representation of urban scenes. However, due to the lack of semantic information, they can hardly be used directly in various real-world applications, such as urban planning, simulation, and solar potential estimation.

\begin{figure}[!t]
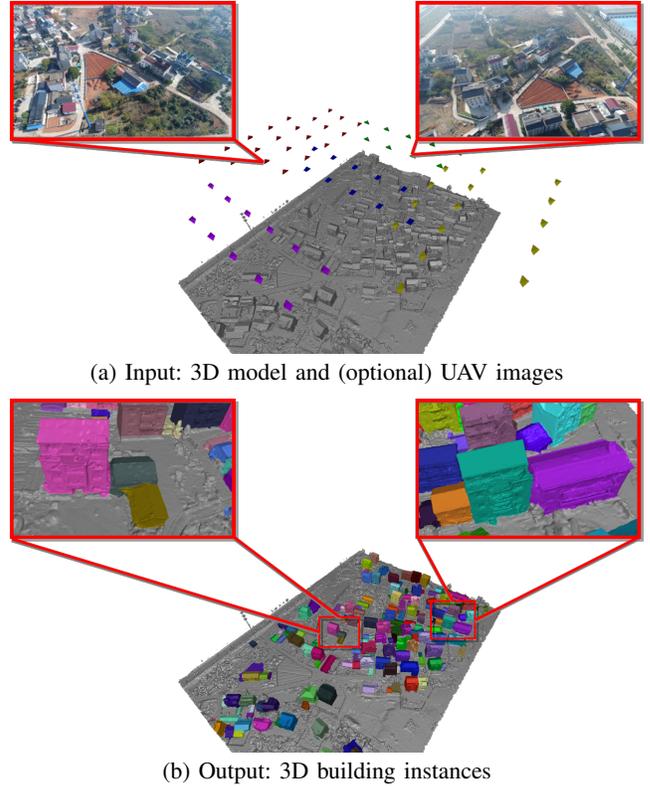

	\centering
    \includegraphics[width=0.95\linewidth]{\imgfile{fig-goal3-1}}\\
    \small{(a) Input: 3D model and (optional) UAV images} \\ \vspace{0.2cm}
    \includegraphics[width=0.95\linewidth]{\imgfile{fig-goal3-2}}\\
    \small{(b) Output: 3D building instances}
	\caption{Instance segmentation of 3D buildings in a large urban scene. The pyramids above the 3D scene model in (a) indicate the position and orientation of the cameras.}
	\label{fig-goal}
\end{figure}

Buildings are the most important part of a city, its segmentation is the core of the semantic analysis of urban scenes. Rather than semantic segmentation, we focus on the instance segmentation of buildings, as it separates different building instances, even if they are attached. Thus, our goal is to segment all building instances in a large 3D urban scene precisely and automatically, as shown in Fig.\ref{fig-goal}.

Recent advances in deep learning have achieved great success in image instance segmentation, but applying these techniques to 3D mesh models is still challenging and has not been sufficiently explored\cjz{, especially for 3D buildings in large-scale urban scenes}. Feature extraction of 3D data is the key to applying deep learning to the 3D model segmentation. Volumetric-based methods~\cite{riegler2017octnet,maturana2015voxnet}, and point cloud-based methods~\cite{qi2017pointnet,qi2017pointnet++} are limited by memory and computing power, thus are mainly used for the segmentation of relatively small indoor scenes~\cite{yi2019gspn,wang2018sgpn,hou20193d,wang2019associatively,pham2019jsis3d,liu2019masc}. Besides, the lack of annotated 3D instance segmentation dataset for outdoor scenes also hinders the application of deep learning on instance segmentation of 3D urban scenes.

\cjz{Instead of directly segmenting 3D models, segmenting images first and projecting them to the 3D models is a potential alternative, as it can utilize powerful neural networks for image segmentation. Orthophoto maps could be the first candidate, due to their unified projection directions. However, buildings in orthophoto maps have severe self-occlusion, e.g. walls cannot be seen. Thus, its segmentation inaccuracy will be conducted to the 3D models during the 2D-3D projection. For this sake, we employ a multi-view 3D segmentation framework in this paper. It first employs existing learning-based 2D instance segmentation to segment roofs in drone images, back-projects roof instance masks to the 3D scenes considering the spatial occlusion, and finally constructs Markov random fields to segment out all buildings from 3D scenes. }

\cjz{Such a multi-view framework for 3D semantic segmentation tasks is straightforward, and a winner-take-all mechanism works well for most semantic segmentation tasks. However, a multi-view framework for 3D instance segmentation is much more challenging, as instances have to keep separating even if they are spatially connected. Projecting instance masks back to the 3D scene without correspondences will lead to instance ambiguities, as 2D instance masks from different views may be inconsistent. Buildings in multi-view images are often partially occluded by attached other buildings or tall trees, which increases ambiguities in mask correspondences.}


In this paper, we propose a novel instance mask clustering method to build mask correspondences among multi-view UAV images. \cjz{It solves the issue of instance ambiguities robustly, making our multi-view 3D instance segmentation method outperform the orthophoto-based method.} To improve segmentation accuracy for diversely distributed buildings, we enhance multi-view RGB images to RGBH images by adding an extra channel encoding the height information. To ease the spatial occlusion challenge, we perform instance segmentation of roofs instead of entire buildings, since roofs have higher visibility than other parts of buildings.

\cjz{Though our method incorporates existing image instance segmentation techniques, it includes the following core contributions:}
\begin{itemize}
	\item A multi-view instance segmentation framework that segments 3D buildings in large urban scenes efficiently and precisely;
	
	\item An occlusion-aware clustering method for instance masks, which robustly eliminates ambiguities in mask correspondences among multi-view images;
	
	\item A benchmark dataset \datasetname\ for instance segmentation evaluation of 3D buildings in large urban scenes, which consists of pixel-level instance annotation for both UAV images and 3D urban models.
	
\end{itemize}

\label{sec:intro}

\section{Related Work}
There is a large volume of research in instance segmentation for various data sources. In this section, we review the existing work of instance segmentation for common images, aerial images, and 3D data.

\subsection{Common image instance segmentation}
Existing instance segmentation methods for natural images can be classified into two categories: object detection-based approaches and metric learning-based ones.

\noindent\textbf{Object-detection based approaches}.
Object detection-based approaches work in a top-down manner and highly depend on object detection or proposal. R-CNN first introduces CNN in the field of object detection~\cite{girshick2014rich}. To improve computational efficiency, Fast R-CNN proposes an improved SPP (Spatial Pyramid Pooling)~\cite{he2015spatial} structure named RoI pooling~\cite{girshick2015fast}, and Faster R-CNN uses region proposal networks instead of selective searching to extract object candidates, which forms an efficient end-to-end network for object detection~\cite{ren2015faster}.

Based on Faster R-CNN~\cite{ren2015faster}, Mask R-CNN combines with FPN~\cite{lin2017feature} to detect objects with different sizes and uses RoIAlign instead of RoI pooling to form a simple, flexible, and effective instance segmentation network~\cite{he2017mask}. Recently, mask scoring R-CNN uses mask scores to improve the category scores used in Mask R-CNN~\cite{huang2019mask}. PANet uses a bottom-up annotation structure to shorten the information path and enhances the feature pyramid with accurate localization signals existing at low levels~\cite{liu2018path}. HTC combines detection and segmentation with a multi-task and multi-stage hybrid cascade structure~\cite{chen2019hybrid}. \xyh{Swin Transformer~\cite{2021Swin} proposes a hierarchical Transformer whose representation is computed with shifted windows. Its hierarchical architecture has flexibility at various scales and linear computational complexity concerning image size, which helps it to achieve outstanding segmentation performance. }

\noindent\textbf{Metric learning-based approaches}.
Many other dense instance segmentation methods are based on metric learning~\cite{de2017semantic}. These methods work in a bottom-up manner, generate embedding features~\cite{fathi2017semantic,novotny2018semi} for each pixel and use post-processing methods such as clustering~\cite{liang2017proposal,kong2018recurrent} or graph theory~\cite{bai2017deep} to classify these pixels. Inspired by FCIS~\cite{li2017fully} and YOLACT~\cite{bolya2019yolact}, BlendMask uses a blender module to merge top-level coarse instance information with lower-level fine granularities~\cite{chen2020blendmask}.

\subsection{Aerial image segmentation}
In the last decade, instance segmentation methods for aerial images of urban scenes have also been proposed because of the wide applications of aerial images. Montoya et al. use an $\alpha$-shape algorithm to calculate the boundary polygons of building objects, which are further optimized by CRF~\cite{montoya2015semantic}. By combining the CNN backbone with FPN and RNN, Li et al. propose an end-to-end deep neural network to predict polygon outlines of buildings and road topology maps~\cite{li2019topological}. Conv MPN uses GNN (graph neural network)~\cite{battaglia2018relational} to reconstruct the building plan from a single image~\cite{zhang2020conv}.
DARNet employs a polar representation of contours to predict contours that are free of self-intersection and a loss function consisting of a data term, a curvature term, and a balloon term, which not only encourages the predicted contours to match ground truth building boundaries but also prefers low-curvature solutions~\cite{cheng2019darnet}.

\cjz{Besides instance segmentation, many semantic segmentation methods for aerial images have been proposed recently in the remote sensing domain. They are also referred to as remote sensing image classification. Besides single-modality images, researchers in the remote sensing domain are also interested in employing deep learning techniques in the pixel-level classification of multi-modality images, including multispectral ones and hyperspectral ones, which are proven to overcome the challenge of information diversity~\cite{hong2020more}. For instance, Hong et al. introduce graph convolutional networks into hyperspectral image classification in a minibatch fashion~\cite{hong2021graph}, and also propose a new transform-based network that learns locally spectral representations from multiple neighboring bands instead of single bands~\cite{hong2021spectral}. Since multispectral and hyperspectral images require expensive and heavy spectrometers to acquire, RGB images are more common for UAVs. In this paper, height maps are automatically generated and added to  corresponding RGB images respectively. They can not provide as rich information as hyperspectral images, but this geometric information is a very important supplement that can significantly improve segmentation accuracy. }


\subsection{3D instance segmentation}
Unlike images that inherently have a grid structure, the vertices and faces in discrete surfaces (i.e., 3D meshes) do not have regular spatial structures to be directly convoluted. Volumetric methods ease this issue by using a 3D grid representation, which is notoriously expensive in terms of computational efficiency and memory consumption~\cite{maturana2015voxnet,liu2019masc,dai2018scancomplete,wu20153d,wang2017cnn,dai2017shape}.

Various strategies have been proposed to address the memory issue of volumetric methods. For example, OctNet uses an octree structure to avoid unnecessary cells~\cite{riegler2017octnet}, thus reducing memory consumption. PointNet uses T-net and max-pooling to achieve rotation invariance and the capability of handling unordered 3D point clouds. It fuses both local and global features, making it an efficient and effective feature extractor for point cloud data~\cite{qi2017pointnet}. Through point grouping and multi-level feature extraction, PointNet++ can better extract discriminative features for point clouds with uneven density~\cite{qi2017pointnet++}.

\begin{figure*}[!t]
\centering
\includegraphics[width=0.9\linewidth]{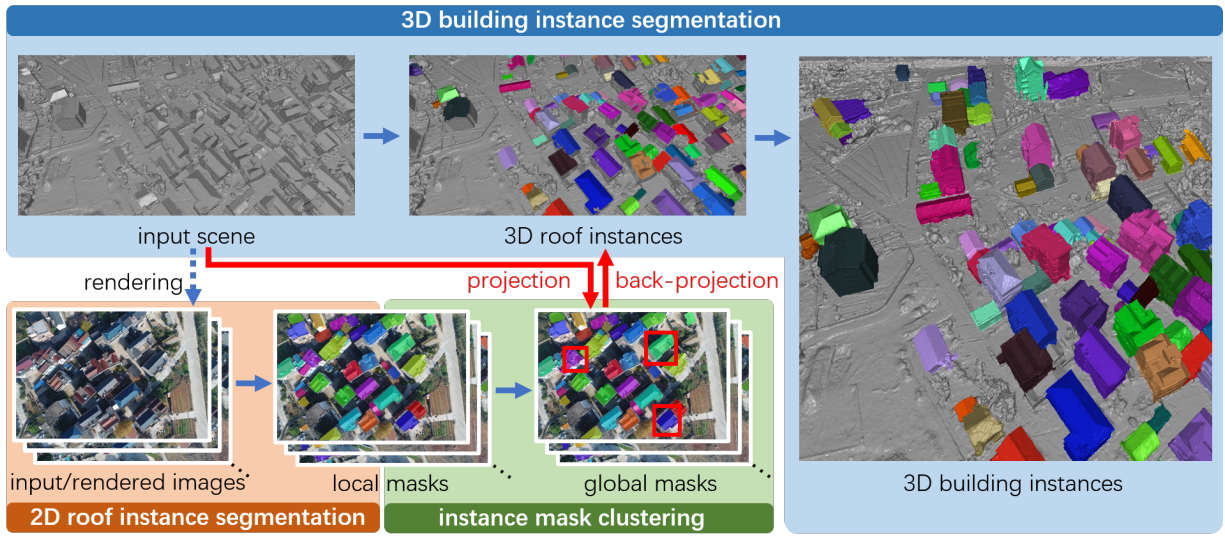}%
\caption{An overview of the proposed method. Our method takes a 3D urban scene and optionally multi-view UAV images as input and segments all 3D building instances as results.
It contains three major steps: 2D roof instance segmentation, instance mask clustering, and 3D building instance segmentation. The multi-view images are not obligatory, as they can also be generated by the rendering of the input 3D scene with textures (noted by the dotted arrow in the figure).
The red rectangles highlight a few global masks selected by our clustering method.
The projection and back-projection operations noted by the red arrows in the figure contribute to both instance mask clustering and the occlusion-aware 3D roof segmentation.}
\label{fig-pipeline}
\end{figure*}

Based on features extracted by PointNet, PointNet++, and PointCNN~\cite{li2018pointcnn}, many 3D instance segmentation methods for point clouds have also been proposed. SGPN predicts the instances by learning the similarity matrix between point clouds. However, the size of its similarity matrix tends to explode as the number of points increases~\cite{wang2018sgpn}. GSPN extends the structure of Mask R-CNN (that was originally developed for images) to process 3D data~\cite{yi2019gspn}.
JSNet~\cite{zhao2020jsnet} and ASIS~\cite{wang2019associatively} both learn the instance embedding space and combine semantic features and instance features of the point clouds to jointly improve the accuracy of semantic segmentation and instance segmentation.

Though great progress has been made for instance segmentation of indoor scenes. However, existing methods are designed for processing point cloud data. It is still a challenge to extend these methods to outdoor scenes without sufficient annotated data and generalize them to handle the fast accumulation of urban models in the form of meshes. In this work, we establish a 3D instance segmentation dataset for urban scenes and propose the first framework for 3D instance segmentation of buildings from urban MVS meshes.

\label{sec:related}

\section{Methodology}

Compared to the lower parts of buildings that are more likely to be occluded by the nearby buildings and trees, building roofs usually have better visibility in aerial imaging. This observation motivates us to approach instance segmentation of entire buildings by looking into the segmentation of building roofs. In contrast to the previous work directly segmenting entire buildings~\cite{li2016reconstructing}, we perform roof segmentation by using a deep neural network. This strategy significantly improves the accuracy of the segmentation stage and simplifies the manual annotation in the data preparation stage.

One characteristic of our method is the hybrid process of 2D images and the corresponding 3D meshes, in which spatial occlusion is fully considered in processing the two distinctive data sources. Fig.~\ref{fig-pipeline} shows the proposed multi-view 3D instance segmentation framework that consists of three major steps:

\begin{enumerate}
	\item \textbf{2D roof instance segmentation}. Roofs in multi-view images are automatically segmented by an instance segmentation neural network that is fine-tuned using our RGBH imagery dataset.
	
	\item  \textbf{Instance mask clustering}. An occlusion-aware clustering method for roof instance masks is exploited, which correlates instance masks from multi-view images to eliminate ambiguities. The mask clustering is the core of our method, which projects the 3D urban scene to the image space to measure the spatial overlap between arbitrary pairs of instance masks.
	
	\item  \textbf{3D building instance segmentation}. The clustered masks are projected back to the 3D space to segment 3D roof instances and the entire buildings are segmented in the end through an MRF optimization.
\end{enumerate}

In the following part of this section, we introduce these three major steps, the benchmark dataset, and the implementation details.

\subsection{2D Roof instance segmentation}
\label{sec-sub-2d-roof-instance-segmentation}

The challenge in building instance segmentation lies in that the roofs of adjacent (or even attached) buildings may have very similar appearances despite the difference in the height of the buildings. In this work, we take advantage of the complementary characteristics of the images and the 3D model of the scene by enhancing each RGB aerial image to an RGBH image that provides additional geometric cues. Specifically, we render a heightmap for each drone image using the 3D urban models reconstructed from the drone images and camera parameters. We add an additional channel encoding the height information to the drone images to obtain RGBH images, in which the height values are separately normalized for each image. With the RGBH images, we apply Swin Transformer~\cite{2021Swin} to segment roof instances automatically.

\begin{figure*}[!t]
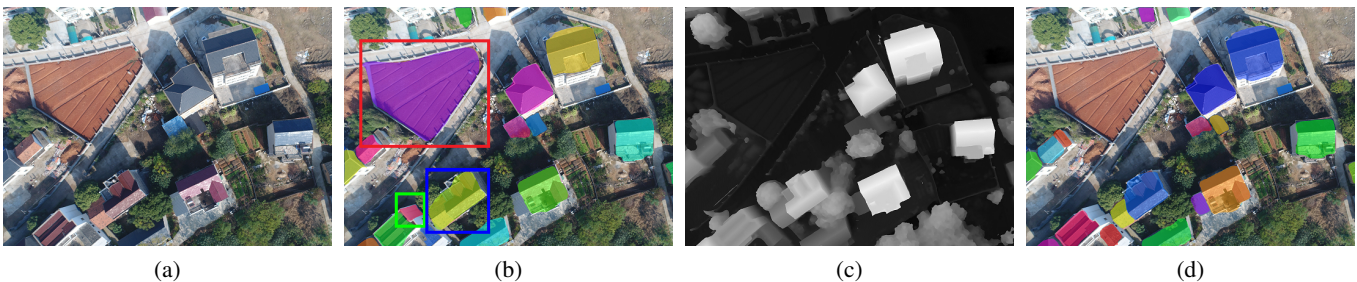

\centering
\begin{tabular}{@{}c@{\hspace*{0.01\linewidth}}c@{\hspace*{0.01\linewidth}}c@{\hspace*{0.01\linewidth}}c@{}}
\includegraphics[width=0.24\linewidth]{\imgfile{seg-2D-org}} &
\includegraphics[width=0.24\linewidth]{\imgfile{seg-2D-RGB}} &
\includegraphics[width=0.24\linewidth]{\imgfile{seg-2D-height}} &
\includegraphics[width=0.24\linewidth]{\imgfile{seg-2D-RGBH}} \\
{\small{(a)}} &
{\small{(b)}} &
{\small{(c)}} &
{\small{(d)}}
\end{tabular}
\caption{Comparison of the segmentation results without and with height information. (a) An input test drone image. (b) The segmentation result using only the RGB image. The rectangles highlight three misclassified regions, i.e., a vegetable field (in red) and a wall (in green) of a building are segmented as roofs, and two roofs of attached buildings (in blue) are not separated. (c) The heightmap obtained by rendering the scene using the 3D model and the camera parameters. (d) The segmentation result using both the RGB image and the heightmap, where the vegetable field, wall, and roofs are all correctly separated.}
\label{fig-seg-image}
\end{figure*}

According to our quantitative evaluation on the benchmark dataset, the AP (average precision)~\cite{lin2014microsoft} of the segmentation of roof instances on RGBH images reaches 0.582, which is significantly higher compared to 0.563 achieved on RGB aerial images. This demonstrates the advantage of height information on the roof instance segmentation. A visual comparison is shown in Fig.~\ref{fig-seg-image}. Ground objects like trees and vegetable fields are successfully separated from buildings, even though some of them have visually indistinguishable textures from building roofs. The Swin Transformer neural network also computes a probability for each instance mask to represent its prediction confidence. To avoid low-confident masks, we only use roof instance masks whose prediction confidence is higher than 70\%.

\subsection{Instance mask clustering}
\label{sec-sub-ins-mask-clustering}
\begin{figure*}
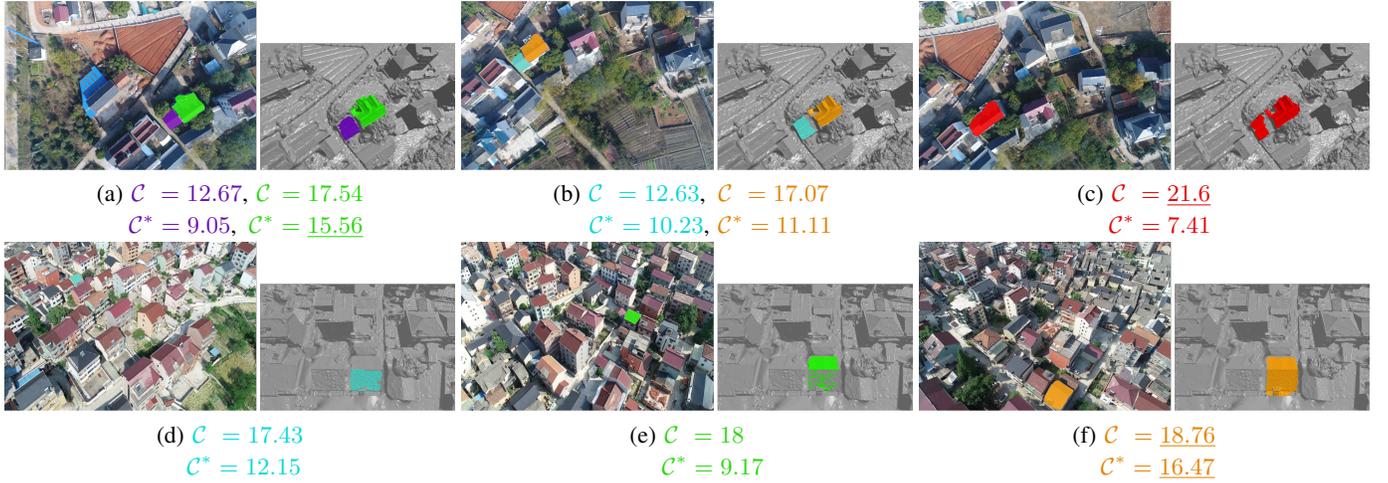

\centering
\begin{tabular}{@{}c@{\hspace*{0.005\linewidth}}c
                @{\hspace*{0.005\linewidth}}c@{}}
\includegraphics[width=0.33\linewidth]{\imgfile{fig-cluster-a}} &
\includegraphics[width=0.33\linewidth]{\imgfile{fig-cluster-b}} &
\includegraphics[width=0.33\linewidth]{\imgfile{fig-cluster-c}} \\
{\small{(a)} $\textcolor[rgb]{0.40,0.05,0.70}{\mathcal{C}\,\,=12.67}$,  $\textcolor[rgb]{0.15,0.85,0.04}{\mathcal{C}\,=17.54}$}&
{\small{\,\,(b)} $\textcolor[rgb]{0.00,0.85,0.85}{\mathcal{C}\,\,=12.63}$, $\textcolor[rgb]{0.90,0.50,0.00}{\,\mathcal{C}\,\,=17.07}$}&
{\small{(c)} $\textcolor[rgb]{0.90,0.00,0.00}{\mathcal{C}\,\,=\underline{21.6}}$ }\\
{\small\quad $\textcolor[rgb]{0.40,0.05,0.70}{\,\,\mathcal{C}^*=9.05}$,  $\textcolor[rgb]{0.15,0.85,0.04}{\,\mathcal{C}^*=\underline{15.56}}$}&
{\small\qquad $\textcolor[rgb]{0.00,0.85,0.85}{\mathcal{C}^*=10.23}$, $\textcolor[rgb]{0.90,0.50,0.00}{\mathcal{C}^*=11.11}$}&
{\small\quad $\textcolor[rgb]{0.90,0.00,0.00}{\,\,\mathcal{C}^*=7.41}$ }\\
\includegraphics[width=0.33\linewidth]{\imgfile{fig-cluster-d}} &
\includegraphics[width=0.33\linewidth]{\imgfile{fig-cluster-e}} &
\includegraphics[width=0.33\linewidth]{\imgfile{fig-cluster-f}} \\
{\small{(d)} $\textcolor[rgb]{0.00,0.85,0.85}{\mathcal{C}\,\,=17.43}$ }&
{\small{(e)} $\textcolor[rgb]{0.15,0.85,0.04}{\mathcal{C}\,\,=18}$ }&
{\small{(f)} $\textcolor[rgb]{0.90,0.50,0.00}{\mathcal{C}\,\,=\underline{18.76}}$} \\
{\small\quad $\textcolor[rgb]{0.00,0.85,0.85}{\,\mathcal{C}^*=12.15}$ }&
{\small\qquad $\textcolor[rgb]{0.15,0.85,0.04}{\,\mathcal{C}^*=9.17}$ }&
{\small\quad $\textcolor[rgb]{0.90,0.50,0.00}{\,\mathcal{C}^*=\underline{16.47}}$} \\
\end{tabular}
\caption{Ambiguities in 2D roof instance segmentation. In each row, instance masks are shown in both the drone images (in different views) and the corresponding 3D mesh models (in an identical view). Top row: two roofs are separated in (a) and (b) but incorrectly mixed together in (c). Bottom row: (d) and (e) only cover a small part of the same roof, while (f) covers the entire roof. Original mask confidence values (denoted by $\mathcal{C}$) and improved mask confidence values  (denoted by $\mathcal{C}^*$) are given in the sub-captions. The underlined numbers are the highest confidence values in each row. }
\label{fig-cluster}
\end{figure*}

After the roof instance segmentation, we obtain a set of roof instance masks from multi-view images, where multiple masks may correspond to the same roof.
Since roofs of the same building in multiple views have been segmented independently, the correspondences between roof instance masks are not known. This results in the number of instance masks being much larger than the number of roofs in the scene.
To establish the correspondences between the masks from multi-view images, we refer to 3D roof instances by back-projecting the 2D roof masks onto the 3D model of the scene using the camera parameters. However, identifying masks that correspond to the same roof is challenging due to two main reasons:
First, the 2D instance segmentation may have errors due to the limited capability of the neural network and the complex structure of the building roofs, as shown in the first row of Fig. \ref{fig-cluster}. Second, the instance masks from different views are ambiguous due to different levels of spatial occlusion, as shown in the second row of Fig. \ref{fig-cluster}.

To tackle these two challenges, we propose an instance mask clustering method that divides instance masks into different groups such that each group corresponds to a unique roof instance of an individual building. Representative masks are first selected from the segmented instance masks, and the remaining masks are merged with them according to mask similarity measures.
For clarity, we refer to all roof instance masks in multi-view images as local masks, while the representative masks selected for clustering as global masks, as they represent unique building roofs across different images.

We first build a similarity matrix $\mathcal{M}$ to measure the spatial overlap for each pair of local masks. Based on $\mathcal{M}$, a mask with confidence value $\mathcal{C}$ to be selected as a global mask is computed for each local mask.
Finally, all local masks are sorted in descending order to select reliable global masks and are clustered into groups according to their similarities with the global masks. Note that each mask group contains only one global mask. We establish the mapping between all local masks and global masks. In the following, we elaborate on these steps in detail.

\noindent\textbf{Occlusion-aware mask similarity.}
To measure the spatial overlap of two masks, we project 3D mesh triangles to the image using the camera parameters. We render a depth map with the GPU acceleration for each multi-view image and employ a depth test to check the visibility for all the vertices. For the $i$th local mask, we record a set of triangles $S_i$ whose centers are projected within this local mask region. A similarity matrix $\mathcal{M}_{n\times n}$ is then computed to quantify the spatial overlap between every pair of local masks, where $n$ is the number of all local masks. The similarity element $m_{ij}$ measures the intersection over union (i.e., \emph{IoU}) between the $i$th and the $j$th local masks, i.e.,

\begin{equation}
m_{ij} = A(S_i \cap S_j) / A(S_i \cup S_j),
\label{equ-matrix}
\end{equation}
where $A(S)$ is the surface area of the triangles in the set $S$. $\mathcal{M}_{n\times n}$ is a symmetric matrix as $m_{ij} = m_{ji}$.

\noindent\textbf{Mask confidence.}
Generally, an ideal global mask should have the most overlap with local masks that correspond to the same roof and have the least overlap with local masks that correspond to roofs of different buildings. To select such global masks, we estimate a confidence value $\mathcal{C}$ for each local mask to evaluate the overall overlap with all other local masks in the scene. It is calculated as the sum of similarity elements on the $i$th row of the similarity matrix $\mathcal{M}$:
\begin{equation}
\mathcal{C}_i = \mathcal{P}_i \cdot \sum_{j=1}^n  \mathcal{P}_j \cdot m_{ij},
\label{equ-ratio}
\end{equation}

\noindent where $\mathcal{P}_i$ is the probability value produced by the Swin Transformer neural network, $\mathcal{C}_i$ sums up the probability-weighted similarity elements of local masks. It makes sense because local masks with higher prediction confidences are more likely to be global masks.

However, there is still one drawback in Equation (\ref{equ-ratio}): local masks with large areas may suppress smaller ones because they likely overlap more with other local masks, and thus they obtain larger mask confidence values. Local masks with larger areas are not always the ideal global masks. The top row of Fig.\ref{fig-cluster} shows such a counter-example. To solve this issue, we define a binary term $\Delta_{ij}$ to avoid such unexpected suppression:

\begin{equation}
\Delta_{ij} = \delta(m_{ij}-\beta),
\label{equ-overlap}
\end{equation}

\noindent where $\delta(\cdot)$ is the delta function:
 \begin{equation}
\delta(x) = \begin{cases}
0,  & \mbox{if }x\ \leq  0 \\
1, & \mbox{if }x\ >  0
\end{cases}.
\label{equ-delta}
\end{equation}

\noindent With this binary term, the mask confidence is updated to:

\begin{equation}
\mathcal{C}^*_i = \mathcal{P}_i \cdot \sum_{j=1}^n \Delta_{ij} \cdot \mathcal{P}_j \cdot m_{ij}.
\label{equ-overlap}
\end{equation}

\noindent $\mathcal{C}^*_i$ sums up the similarity elements $m_{ij}$ whose values are higher than a threshold $\beta \in [0,1]$ for each local mask. When $\beta$ is close to 0, $\Delta_{ij}=1$ for most cases and thus $\mathcal{C}^*_i$ degenerates to $\mathcal{C}_i$. When $\beta$ is close to 1, $\Delta_{ij}$ and $\mathcal{C}^*_i$ are both close to 0, which means that the confidence values of all local masks to be selected as global masks are close to 0. In such a case, the clustering cannot distinguish global masks from local masks, resulting in false clustering in the end. In this work, we set $\beta = 0.5$ in all our experiments, indicating that local masks with similarities (i.e., the ratio of the overlapping area) higher than $\beta$ are considered in the computation of mask confidence values. More details about the evaluation of the parameter $\beta$ can be found in the implementation details in Subsection~\ref{subsec-para-eval}.

\noindent\textbf{Mask clustering.}
One key observation of this work is that local masks with higher confidence values are consistent with other masks and thus should have higher priority to be selected as global masks, as shown in Fig. \ref{fig-cluster}.

Based on the mask confidence, we employ a simple yet efficient order-based mask clustering. We first sort all local masks according to their confidence values $\mathcal{C}^*$ and then traverse them in descending order to select global masks. In the traversing loop, if a local mask has not been marked, we mark it as a new global mask, and other non-marked local masks whose similarities with this global mask are higher than $\beta$ are considered consistent with this global mask, i.e., $\delta(m_{lg}-\beta)=1$ where $l$ and $g$ are the indices of the local mask and this global mask, respectively. If a local mask has been already marked, we traverse to the next local mask until all of them are marked.

With the pre-computation of mask confidence values, the traversal is required only once. For efficiency, we establish a mapping table $\mathbb M$ between all local masks to their corresponding global masks. It is worth noting that even though we cannot guarantee each global mask in $\mathbb M$ corresponds to a building instance in the real scene at this stage, a few false correspondences will not affect the final 3D instance segmentation. This will be explained in Subsection~\ref{sec-sub-3DroofSeg}.

\subsection{3D building instance segmentation}
\label{sec-sub-buildingseg}
\noindent\textbf{3D roof instance segmentation.}
\label{sec-sub-3DroofSeg}
The existing multi-view semantic segmentation framework projects the 2D semantic labels back to the 3D model with the highest probability~\cite{Kundu2014joint,blaha2016large}. For 3D instance segmentation, this framework is not suitable because the correspondences between local masks from multi-view images are unknown. Our mask clustering establishes the correspondences between the local masks from multiple views. We first project each vertex of the 3D model to all images and check its visibility using fields of view and depth maps. Then, using its corresponding local mask index at its projected position in the image, we retrieve its corresponding global mask from the mask mapping table $\mathbb M$.


For a vertex $v$, let $MVI_{v}$ denote the set of multi-view images in which $v$ is visible and $GMI_{p}$ denote the global mask index corresponding to a multi-view image $p$.
In some cases, a vertex $v$ on the 3D surface model may be projected within multiple global masks. We denote the set of these global masks as $\{{GMI}_p|p\in {MVI}_v\}$ and thus ${GMI}_p=-1$ represents the background. From these global masks, the one with the largest quantity of corresponding local masks that were projected to by this vertex is associated with this vertex:
\begin{equation}
{RID}_v = maxCount(\{{GMI}_p|p\in {MVI}_v\}),
\label{equ-3d-vertex-seg}
\end{equation}
where ${RID}_v$ is the roof ID of vertex $v$, and the function $maxCount(S)$ extracts the most occurring element in the set $S$. In case more than one global masks have the maximum count in the set $S$, the global mask with a smaller value of $GMI$ will be chosen, as a smaller $GMI$ value corresponds to a higher confidence of the global mask.

The advantage of determining the roof IDs in this way is that the most confident global mask can be automatically selected, and thus a user does not have to provide a specific number of target clusters (i.e., the number of global masks).
This is because local masks with large errors in 2D roof segmentation are normally divided into groups containing small numbers of local masks. The global masks derived from these local masks usually have wrong predictions and therefore will be ignored in the upcoming 3D roof segmentation step, since only the global mask with the largest quantity of corresponding local masks is selected. Therefore, our 3D roof instance segmentation achieves higher prediction precision than the roof instance segmentation on the multi-view images. Their AP/AP50/AP75 values can be found in Subsection~\ref{sec-sub-2d-roof-instance-segmentation} and Table~\ref{tbl-roof-eva} respectively.

With the roof IDs for all vertices, the segmentation of the 3D model can be easily obtained. Specifically, the roof ID of a triangle face is determined by the majority of the vertices that indicate the same roof ID, i.e.,
\begin{equation}
{RID}_t = maxCount(\{{RID}_v|v\in V_t\}),
\label{equ-3d-triangle-seg}
\end{equation}
where ${RID}_t$ represents the roof ID of triangle $t$, and $V_t$ represents the three vertices of the triangle $t$.

\begin{figure}[!t]
\centering
\includegraphics[width=0.7\linewidth]{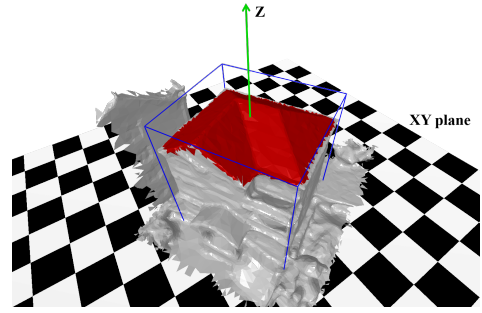}%
\caption{Horizontal Oriented Bounding Box ($HOBB$). The top and bottom faces of the $HOBB$ are horizontal, and the other four faces are oriented according to the PCA orientation estimation.}
\label{fig-HOBB}
\end{figure}

\noindent\textbf{MRF-based 3D building segmentation.}
Based on 3D roof segmentation, the next step is to segment the entire 3D buildings. We first estimate a Horizontal Oriented Bounding Box ($HOBB$) for each roof instance, as shown in Fig.~\ref{fig-HOBB}. To segment an entire building from a 3D scene, we expand the $HOBB$ on its four sides by a certain offset value (4 meters in all of our experiments). The triangles within the expanded $HOBB$ (excluding those that have been labeled by other roof instances) are selected as a candidate building. We denote its triangle set as $T=\{t_i\}$ and its edge set as $E=\{e_{ij}\}$, in which $t_i$ represents the $i$th triangles in $T$, and $e_{ij}$ represents the edge connecting $t_i$ and $t_j$.
We formulate the building segmentation as a foreground/background labeling process that minimizes the following energy function:

\begin{equation}
\psi(l) = \sum_{t_i\in T} \psi_{data}(l_i) + \sum_{e_{ij}\in E} \psi_{smooth}(l_i,l_j),
\label{equ-mrf}
\end{equation}
where $l_i$ denotes the label of triangle $t_i$ given by MRF-based segmentation. $l_i=1$ indicates the foreground (i.e., a building triangle) and $l_i=0$ the background (i.e., a non-building triangle).

\begin{figure*}
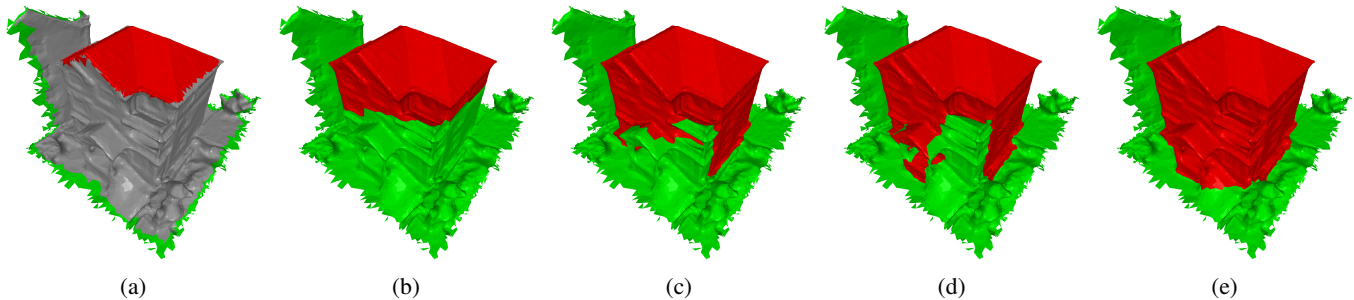

\centering
\begin{tabular}{@{}c@{\hspace*{0.005\linewidth}}c@{\hspace*{0.005\linewidth}}c@{\hspace*{0.005\linewidth}}
c@{\hspace*{0.005\linewidth}}c@{}}
\includegraphics[width=0.195\linewidth]{\imgfile{seg-init}} &
\includegraphics[width=0.195\linewidth]{\imgfile{seg-half}} &
\includegraphics[width=0.195\linewidth]{\imgfile{seg-dis}} &
\includegraphics[width=0.195\linewidth]{\imgfile{seg-nor}} &
\includegraphics[width=0.195\linewidth]{\imgfile{seg-dis-nor}} \\
{\small (a)} &
{\small (b)} &
{\small (c)} &
{\small (d)} &
{\small (e)}
\end{tabular}
\caption{3D building segmentation based on MRF optimization.
	(a) Initial segmentation by the projection of the 2D roof onto the 3D model (foreground in red and background in green).
    (b) Result of direct MRF optimization, where the initial roof segmentation misses the balcony and the gate shelter.
	(c) Our segmentation result without the orientation constraint.
	(d) Our segmentation result without the distance constraints.
	(e) Our segmentation result using both orientation and distance constraints.}
\label{fig-seg-new}
\end{figure*}

The data term $\psi_{data}(l_i)$ represents the penalty of assigning a label $l_i$ to a triangle $t_i$. The 3D roof segmentation provides us a good foreground initialization, we denote its triangle set as $T_f$. We take triangles on the boundary of $T$ as a background initialization, denoted as $T_b$. We further denote the set of other triangles in $T$ as $T_r$. As shown in Fig.~\ref{fig-seg-new}(a), $T_f$, $T_b$, and $T_r$ are visualized in red, green, and gray, respectively.


For triangles in $T_r$, our data term $\psi_{data}(l_i)$ is defined as:
\begin{equation}
\psi_{data}(l_i) = \left\{ \begin{array}{ll}
(1+d_i) +  \theta_i\cdot (1+d_i) & \,\,\,\,\textrm{if}\,\, l_i=1 \\
1 & \,\,\,\,\textrm{if}\,\, l_i=0 \\
\end{array}  \right. .
\label{equ-data-term}
\end{equation}

\begin{figure}[!t]
\centering
\includegraphics[width=0.7\linewidth]{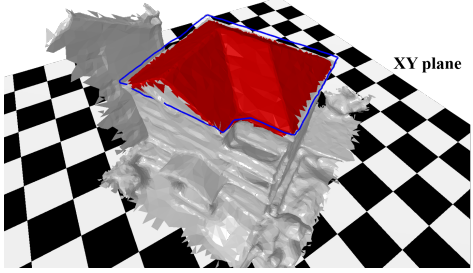}%
\caption{Simplified roof boundary edge $\mathcal{P}$. The blue polygon represents the flat boundary of the roof on the XY plane. To make it easy to observe, we set its Z coordinate to be close to the height of the roof.}
\label{fig-roof-region}
\end{figure}

Before explaining the meaning of $d_i$ and $\theta_i$, we define $\mathcal{P}$, a 2D polygon representing the simplified roof boundary edges, as shown in Fig.~\ref{fig-roof-region}.
We first extract boundary edges of the roof triangles using the Alpha Shape algorithm~\cite{Bernardini97samplingand}. Then, we reduce the dimension of these boundary edges to the 2D horizontal plane by discarding their height and simplifying them through a RANSAC process~\cite{Schnabel2007}. The RANSAC process iteratively merges outline points of roofs to their neighbors if they are approximately collinear. The process stops until no more points can be merged. The simplified roof boundary edges constitute a 2D polygon that is referred to as the roof profile, denoted as $\mathcal{P}$.
For each triangle $t_i \in T_r$, we find a line segment in $\mathcal{P}$ with a minimal distance to the center of $t_i$. We denote this minimal distance as $d_i$, and the cosine of the horizontal angle between the normal of $t_i$ and the direction of this line segment as $\theta_i$. $\theta_i=0$ when the normal of $t_i$ is perpendicular to this line segment, and $\theta_i=1$ when their horizontal angle is $0$. Then $d_i$ is normalized by the maximum distance in the $HOBB$, and we sign the distance as negative if $t_i$ is inside of $\mathcal{P}$. To make the foreground penalty and background penalty comparable on the roof profile (i.e., both penalties equal to 1 when $d_i=0$ and $\theta_i=0$), we use $1+d_i$ instead of $d_i$.

In Equation~\ref{equ-data-term}, the $(1+d_i)$ term is a distance constraint that guarantees only triangles close to $\mathcal{P}$ are taken into the building. The $\theta_i\cdot (1+d_i)$ term is an orientation constraint that guarantees that only triangles having similar orientations with the closest line segment are taken. The $(1+d_i)$ multiplication is to reduce the orientation constraint if the triangles are far away from $\mathcal{P}$.

The smoothness term  $\psi_{smooth}(l_i,l_j)$ penalizes adjacent triangles $t_i$ and $t_j$ being assigned with different labels. We take the cosine angle between normals of adjacent triangles as the penalty for assigning different labels to the adjacent triangle pair, i.e.,

\begin{equation}
\psi_{smooth}(l_i,l_j) = \left\{ \begin{array}{ll}
||n_i\cdot n_j|| & \,\,\,\,\textrm{if}\,\, l_i\neq l_j \\
0 & \,\,\,\,\textrm{if}\,\, l_i=l_j
\end{array} \right. .
\end{equation}
Using the angles between faces, $\psi_{smooth}(l_i,l_j)$ favors segmentation at sharp edges rather than at planar regions.

\subsection{Benchmark dataset}
\label{sec-sub-dataset}

\begin{table}
	\footnotesize
	\renewcommand{\arraystretch}{1.3}
	\centering
	\caption{Statistics on the 3D models of our \datasetname\ dataset.}
	\setlength{\tabcolsep}{1mm}
	\begin{tabular}
		{c|c c c c l}
		\hline
		\textfig{Scene} & \textfig{\#Vertices} & \textfig{\#Triangles} & \tabincell{c}{\textfig{Area}\\ \linespread{-1.5} \textfig{(km$^2$)}} &  \tabincell{c}{\textfig{\#Images}\\ \linespread{-1.5} \textfig{(resolution)}} &  \tabincell{c}{\textfig{\#Buildings}\\ \linespread{-1.5} \textfig{All / Attached }} \\
		\hline
		\#1 & 1.13M & 2.26M & 0.076 & 79 (5472$\times$3648) & \hspace{3mm}185 / 145 \\
		\#2 & 0.87M & 1.72M & 0.097 & 64 (5472$\times$3078) & \hspace{3mm}119 / 40\\
		\#3 & 1.14M & 2.28M & 0.081 & 284 (1916$\times$994) & \hspace{3mm}322 / 232\\
		\#4 & 0.60M & 1.20M & 0.18 & 240 (1536$\times$994) &  \hspace{3mm}266 / 185\\
		\hline
	\end{tabular}
	\label{tbl-result-scene}
\end{table}

We have created a benchmark dataset \datasetname\ that contains annotation for both UAV images and 3D urban scenes simultaneously. To evaluate our 3D instance segmentation method, we annotated 3D roofs and buildings for 4 large 3D urban scenes, which are reconstructed using \textit{Bentley Acute3D ContextCapture}\footnote{https://www.bentley.com/en/products/brands/contextcapture} from UAV images. Table~\ref{tbl-result-scene} shows detailed information about these scenes, and their visualization can be found in Fig.~\ref{fig-result-scene} and the supplementary video. Note that the town is quite crowded, thus about $2/3$ buildings are attached to others, as shown in the last column of Table~\ref{tbl-result-scene}. To facilitate the 3D annotation, we have developed a simple but efficient brush-based annotation tool. Similar to most 2D annotation tools~\cite{Russell2008labelme} which semi-automatically extract pixels of an object by marking the closed boundary polygon of the object, our tool allows a user to segment a 3D building by casually drawing strokes on the building boundaries.

Our \datasetname\ dataset also contains 608 annotated images with high resolutions. They are selected from around 20 thousand images acquired in more than 10 different cities. Some are directly captured by a consumer DJI drone Phantom 4 Pro with different cameras and flight altitudes, others are rendered by 3D models with textures as orthophotos with similar resolutions. There are about 16 thousand buildings in all these images, and their roofs are all manually annotated for the training of our 2D roof instance segmentation neural network. These annotated images are divided into 2 groups, 524 images for training and 84 images for validation.

For accuracy and efficiency consideration, we modified the LabelMe~\cite{Russell2008labelme} toolkit to visualize the corresponding heightmap window alongside the color image window. By synchronizing the annotation on both the image window and the heightmap window, volunteers can freely annotate on either of them. Based on our time recording of volunteer annotation, this double-window strategy saves the volunteers more than half of the time.

\cjz{Our \datasetname\ dataset contains building instance annotation for both 3D urban scenes and UAV images simultaneously, which makes it unique. Most of existing 3D datasets are designed for semantic segmentation, such as Vaihingen3D~\cite{NIEMEYER2014Contextual}, Swiss3DCities~\cite{Can2021Semantic}, Hessigheim3D~\cite{KOLLE2021Michael}, and SUM~\cite{gao2021sum}. The most related work about 3D instance segmentation dataset is the Urban Drone Dataset (UDD)~\cite{chen2018large} and UrbanScene3D~\cite{UrbanScene3D}. UDD evaluates their 3D segmentation accuracy by projecting them to drone images, thus cannot be regarded as a 3D dataset. UrbanScene3D does not contain corresponding UAV images, has only 485 annotated buildings and very few buildings are attached to others. As shown in Table~\ref{tbl-result-scene}, our \datasetname\ dataset has 892 annotated buildings, of which 602 are attached to others. In such crowded urban scenes, instance segmentation has a more significant advantage over semantic segmentation.}

\cjz{Compared to existing natural image datasets, such as MSCOCO~\cite{lin2014microsoft}, Cityscapes~\cite{cordts2016cityscapes}, and PASCAL VOC~\cite{everingham2010pascal}, our annotation on UAV images focuses on roof instances for UAV images. Compared to other remote sensing imagery datasets, such as SpaceNet~\cite{van2018spacenet} and xView~\cite{lam2018xview}, our UAV images have higher resolution and more viewing angles, thus are an important supplement to existing datasets. Though there are also many drone datasets in public, such as VisDrone~\cite{zhu2021detection} and DOTA~\cite{Xia2018Dota}, few of them focus on instance segmentation at the pixel level. With the increasing popularity of low-altitude UAV capturing devices, we believe that our dataset will play an important role in applications such as urban planning, smart cities, and other related fields.}

\subsection{Implementation details}
\label{subsec-imple-detail}

\textbf{2D roof segmentation.} \cjz{We use the Pytorch implementation of Swin Transformer released by~\cite{liu2021Swin} for roof instance segmentation from images. The manually labeled images (see Subsection~\ref{sec-sub-dataset}) were used to fine-tune the Swin Transformer network trained previously using the COCO dataset~\cite{lin2014microsoft}. For 3D urban scenes that have corresponding UAV images, we directly segment these images. For 3D urban scenes that do not have corresponding UAV images, multi-view images are rendered using these 3D scenes from a set of different viewpoints sampled randomly at 70 meters higher than the average height of these scenes. At each viewpoint, 5 virtual cameras facing down, front, back, left, and right are placed.}


\textbf{MRF-based segmentation.}
With the aforementioned MRF setting, we treat Equation~\ref{equ-mrf} as a classical max-flow/min-cut optimization and solve it using the graph cut algorithm~\cite{Boykov2001Interactive}. The MRF-based segmentation result is shown in Fig.~\ref{fig-seg-new} (e). To prove the validity of Equation~\ref{equ-data-term}, we simply set the penalties to 1 for both foreground and background in Equation~\ref{equ-data-term}, as the triangles in $T_r$ have no explicit foreground/background priorities. As a result, it does not produce correct segmentation as expected, as shown in Fig.~\ref{fig-seg-new} (b). Comparisons in Fig.~\ref{fig-seg-new} (c) and (d) show that the introduction of the orientation and the distance constraint overcomes the interference of structural variations and noises in the MRF optimization, thus improving the 3D building segmentation.

\label{sec:method}

\section{Results and Discussion}

We have evaluated our method with both drone images and virtually rendered images. All experiments were carried out on a machine with an Intel Core i7 processor, 32 GB memory, and an NVidia GeForce 1080 GPU.

\subsection{Qualitative results}
\label{subsec-qualitative}
\begin{figure*}
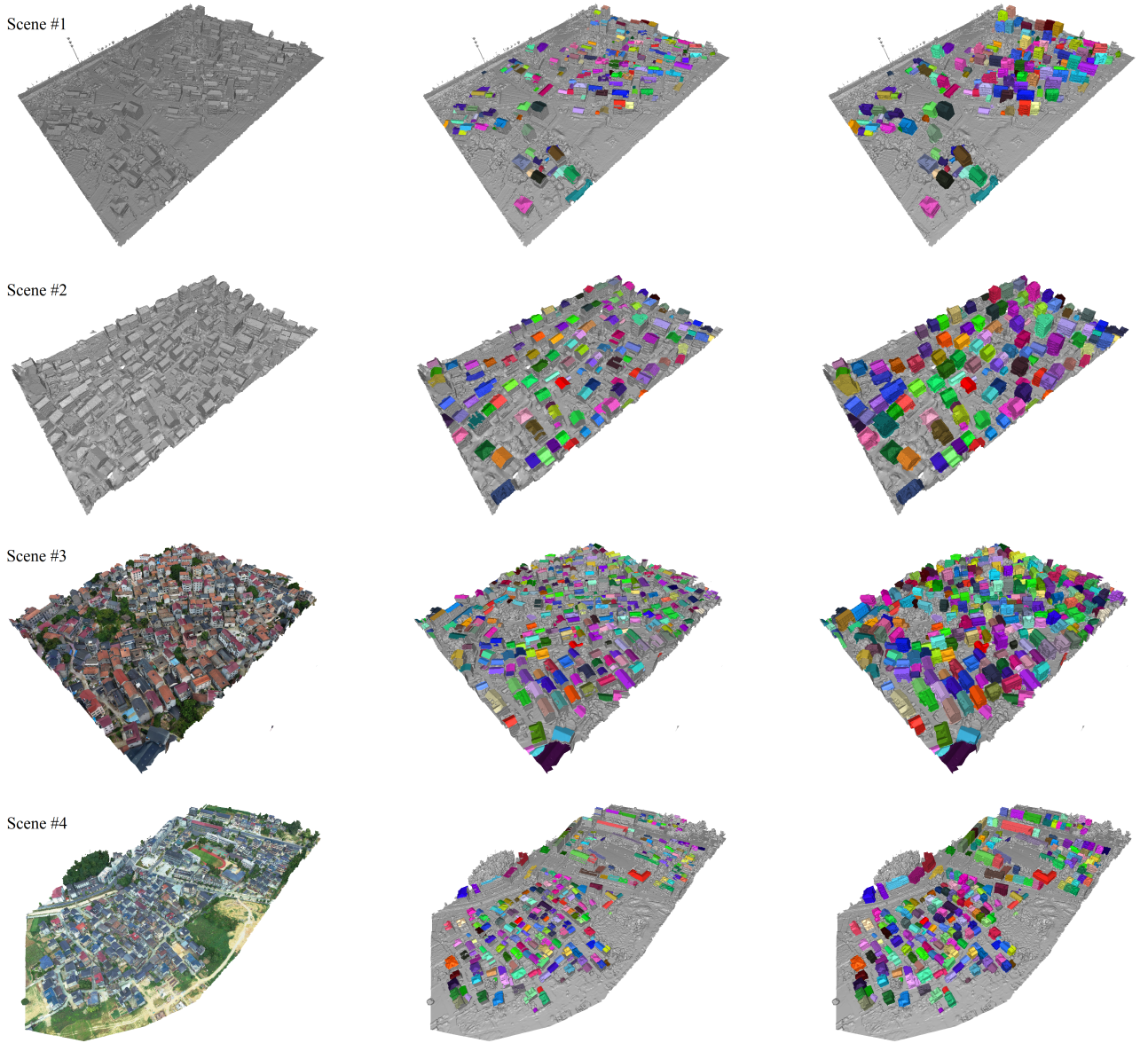

\centering
\begin{tabular}{@{}c@{\hspace*{0.005\linewidth}}c@{\hspace*{0.005\linewidth}}c@{}}
\includegraphics[width=0.33\linewidth]{\imgfile{fig-scene1-init}} &
\includegraphics[width=0.33\linewidth]{\imgfile{fig-scene1-roof}} &
\includegraphics[width=0.33\linewidth]{\imgfile{fig-scene1-building}} \\
\includegraphics[width=0.33\linewidth]{\imgfile{fig-scene2-init}} &
\includegraphics[width=0.33\linewidth]{\imgfile{fig-scene2-roof}} &
\includegraphics[width=0.33\linewidth]{\imgfile{fig-scene2-building}} \\
\includegraphics[width=0.33\linewidth]{\imgfile{fig-scene3-init}} &
\includegraphics[width=0.33\linewidth]{\imgfile{fig-scene3-roof}} &
\includegraphics[width=0.33\linewidth]{\imgfile{fig-scene3-building}} \\
\includegraphics[width=0.33\linewidth]{\imgfile{fig-scene4-init}} &
\includegraphics[width=0.33\linewidth]{\imgfile{fig-scene4-roof}} &
\includegraphics[width=0.33\linewidth]{\imgfile{fig-scene4-building}} \\
(a) Initial scene models & (b) 3D Roof segmentation results & (c) 3D Building segmentation results\\
\end{tabular}
\caption{\cjz{Building instance segmentation results of four scenes. The two scenes on the top have UAV images with camera parameters, while the scenes \#3 and \#4 in the bottom do not have UAV images, for which we render images from multiple viewpoints instead.}}
\label{fig-result-scene}
\end{figure*}

We tested our method on four scenes (see Fig.~\ref{fig-result-scene}), for which the statistics are shown in Table~\ref{tbl-result-scene}.
Scene \#1 and Scene \#2 have UAV images and the corresponding camera parameters. Scene \#3 and Scene \#4 do not have UAV images, we rendered images from multiple viewpoints instead, as explained in the subsection~\ref{subsec-imple-detail}. The roof instance segmentation results are shown in Fig.~\ref{fig-result-scene} (middle column) and the 3D building instance segmentation results are shown in Fig.~\ref{fig-result-scene} (right column).
We can see that our approach successfully segmented most buildings, even if they are dense, varying in style, and attached.

\begin{table*}
	\scriptsize
	\renewcommand{\arraystretch}{1.3}
	\centering
	\caption{\cjz{Comparison of four different clustering methods on 3D \textit{roof} instance segmentation using Swin Transformer.}}
	\setlength{\tabcolsep}{1mm}
	\linespread{1.3}
	\begin{tabular}
		{c|c c c|c c c|c c c|c c c|c c c|c c c}
		\hline
		\multirow{2}{*}{Scene} & \multicolumn{3}{c|}{Spectral with RGB ($K_1$)} & \multicolumn{3}{c|}{Spectral with RGBH ($K_1$)} & \multicolumn{3}{c|}{Spectral with RGB ($K_2$)} &\multicolumn{3}{c|}{Spectral with RGBH ($K_2$)} & \multicolumn{3}{c|}{Ours with RGB} & \multicolumn{3}{c}{Ours with RGBH} \\
		& AP & AP50 & AP75 & AP & AP50 & AP75 & AP & AP50 & AP75 & AP & AP50 & AP75 & AP & AP50 & AP75 & AP & AP50 & AP75 \\
		\hline
		\#1 & 0.5858 & 0.7841 & 0.5966 & 0.6551 & 0.8239 & 0.7045 & 0.5989 & 0.7784 & 0.6364 & 0.6665 & 0.85227 & 0.7159 & 0.6455 & 0.8409 & 0.6818 &\textbf{0.7114} & \textbf{0.8693} & \textbf{0.7727} \\
		\#2 & 0.5329 & 0.6973 & 0.5502 & 0.4579 & 0.6446 & 0.4463 & 0.5436 & 0.7021 & 0.5633 & 0.5826 & 0.8347 & 0.5785 & 0.5651 & 0.7190 & 0.5880 & \textbf{0.6610} & \textbf{0.8395} & \textbf{0.6707} \\
		\#3 & 0.5778 & 0.7663 & 0.6095 & 0.5501 & 0.7284 & 0.6068 & 0.5900 & 0.8162 & 0.6482 & 0.5966 & 0.8398 & 0.6396 & 0.6179 & 0.8476 & 0.6604 & \textbf{0.6434} & \textbf{0.9072} & \textbf{0.6618} \\
		\#4 & 0.4447 & 0.7444 & 0.4398 & 0.4526 & 0.7256 & 0.4549 & 0.4583 & 0.8008 & 0.4323 & 0.4635 & 0.7820 & 0.4474 & 0.5011 & 0.8383 & 0.4962 & \textbf{0.5248} & \textbf{0.8459} & \textbf{0.5301} \\
		\hline
	\end{tabular}
	\label{tbl-roof-eva}
\end{table*}

\subsection{Ablation analysis}
We have evaluated our results on the 3D mesh models in the \datasetname\ benchmark dataset. We first computed the \emph{IoUs} between predictions and the ground truth based on the area of the mesh triangles. Then we used the commonly used
instance-level evaluation metrics, namely AP, AP50, and AP75, in all evaluations~\cite{lin2014microsoft}, where AP50/AP75 indicates the average precision when \emph{IoU} threshold is set to 0.5/0.75, and AP is averaged over 10 \emph{IoU} thresholds of 0.5:0.05:0.95.


\noindent\textbf{Height information.}
As demonstrated in Fig.~\ref{fig-seg-image}, using RGBH images significantly improves the 2D roof instance segmentation. To understand how the height information improves 3D roof instance segmentation and its effects on 3D building instance segmentation, we have conducted a comparison on the four large 3D scenes from \datasetname\ both with and without height information, using two different clustering methods, i.e., the spectral clustering and our method. The results of the comparison are given in Table~\ref{tbl-roof-eva} and Table~\ref{tbl-building-eva}.
These comparisons have revealed that our method using RGBH images achieves higher accuracy than the one without height information (i.e., using RGB images). This is because the additional heightmaps provide spatial information of the urban scenes to the neural network model, which makes the roofs and buildings more distinguishable.

\noindent\textbf{The multi-view framework.} The 2D segmentation in our method is applied to multi-view images, rather than orthophoto maps. To understand the advantages of this multi-view framework, we have also implemented an orthophoto-based instance segmentation solution for comparison. We first generated the orthophoto maps and their corresponding heightmaps of the same scenes through a render-to-texture technique. These two types of maps were combined to form the orthophoto RGBH maps, which were then used to detect the 2D roof instances. Since there was almost no occlusion for the roofs in the orthophoto maps, mask clustering was not necessary and we directly applied the 3D building segmentation to produce the final results. The statistics of the results are reported in Table~\ref{tbl-multi-ortho-eva}.

From this comparison, we can see that our multi-view method achieves higher AP than the orthophoto-based method. \cjz{There are three main reasons for this improvement. The first advantage of using multi-view images lies in the mask clustering that integrates multiple segmentation results of the identical roofs from different viewpoints. This mask clustering process can be regarded as a cross-correction process, thus eliminating most of the incorrect instance masks from individual multi-view images. Secondly, it is very difficult to separate two roofs if they are attached and have similar textures using orthophoto maps. In contrast, our multi-view framework has much more information from building walls, thus achieving more accurate segmentation. Finally, the original drone images usually have high resolution without texture distortion, but orthograph maps may have these drawbacks due to the texture mapping and synthesis on the 3D meshes.}

\noindent\textbf{Mask clustering.}
The instance mask clustering is crucial to overcome the ambiguities in the multi-view instance masks. Many clustering methods are available in the literature, such as Mean-shift~\cite{comaniciu2002mean} and K-means~\cite{kanungo2002efficient}, but none of them is suitable for our multi-view scenario. These clustering methods require computing the mean of features, which heavily relies on a good feature extractor. For example, the position, size, and mean color of masks cannot accurately describe their features. The other difficulty of these clustering methods is choosing the optimal values of parameters, such as the kernel function for Mean-shift, and the K value for K-means. Note that our clustering method does not require specifying the number of target clusters, which is the core advantage of our method.

What is more important is that it is still unknown how to design comparable features for masks from different views, especially when they have a large variation among different viewpoints. Compared with viewpoint-variant features, the spatial overlap between masks can be calculated accurately. Therefore, we directly calculate a similarity matrix using the spatial overlap between masks, rather than their features.

\begin{figure*}
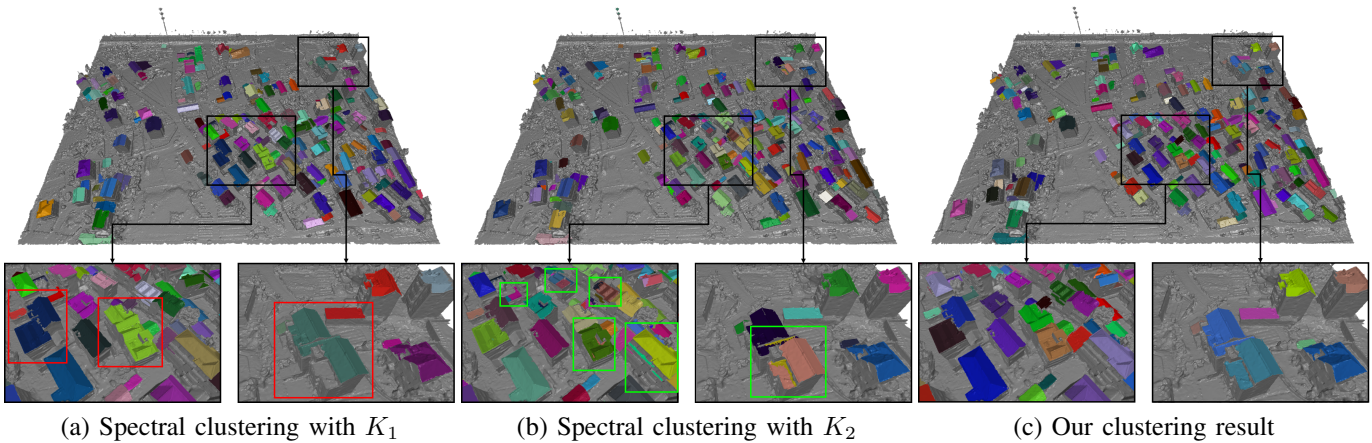

\centering
\begin{tabular}{@{}c@{\hspace*{0.005\linewidth}}c@{\hspace*{0.005\linewidth}}c@{}}
\includegraphics[width=0.33\linewidth]{\imgfile{fig-result-compare-spectral1}} &
\includegraphics[width=0.33\linewidth]{\imgfile{fig-result-compare-spectral2}} &
\includegraphics[width=0.33\linewidth]{\imgfile{fig-result-compare-our}} \\
(a) Spectral clustering with $K_1$  & (b) Spectral clustering with $K_2$ & (c) Our clustering result\\
\end{tabular}
\caption{\cjz{Comparison of roof mask clustering of three methods on Scene \#1. (a) Spectral clustering using the ground truth number of roof targets $K_1$ tends to under-segment (red rectangles) some roofs. (b) Spectral clustering using the number of global masks $K_2$ estimated by our method tends to over-segment (green rectangles) some roofs. (c) Our clustering method achieves more precise roof instance segmentation without specifying the target number.}
}

\label{fig-result-compare}
\end{figure*}

\begin{table*}
\scriptsize
\renewcommand{\arraystretch}{1.3}
\centering
\caption{\cjz{Comparison of four different clustering methods on 3D \textit{building} instance segmentation using Swin Transformer.}}
\setlength{\tabcolsep}{1mm}
\linespread{1.3}
\begin{tabular}
{c|c c c|c c c|c c c|c c c|c c c|c c c}
\hline
 \multirow{2}{*}{Scene} & \multicolumn{3}{c|}{Spectral with RGB ($K_1$)} & \multicolumn{3}{c|}{Spectral with RGBH ($K_1$)} & \multicolumn{3}{c|}{Spectral with RGB ($K_2$)} & \multicolumn{3}{c|}{Spectral with RGBH ($K_2$)} & \multicolumn{3}{c|}{Ours with RGB} & \multicolumn{3}{c}{Ours with RGBH}\\
& AP & AP50 & AP75 & AP & AP50 & AP75 & AP & AP50 & AP75 & AP & AP50 & AP75 & AP & AP50 & AP75 & AP & AP50 & AP75 \\
\hline
\#1 & 0.5881 & 0.7784 & 0.6054 & 0.6362 & 0.8162 & 0.6973 & 0.6184 & 0.8216 & 0.6270 & 0.6843 & 0.8703 & 0.7351 & 0.6486 & 0.8324 & 0.6649 & \textbf{0.7130} & \textbf{0.8919} & \textbf{0.7730} \\
\#2 & 0.5311 & 0.7059 & 0.5210 & 0.4739 & 0.6302 & 0.4705 & 0.5615 & 0.7699 & 0.5666 & 0.5798 & 0.8235 & 0.5798 & 0.5709 & 0.7479 & 0.5966 & \textbf{0.6655} & \textbf{0.8403} & \textbf{0.6807} \\
\#3 & 0.4373 & 0.6429 & 0.4658 & 0.5369 & 0.7764 & 0.5745 & 0.5469 & 0.8168 & 0.5590 & 0.5730 & 0.8354 & 0.5994 & 0.6357 & 0.8882 & 0.6925 & \textbf{0.6671} & \textbf{0.9286} & \textbf{0.7174} \\
\#4 & 0.5739 & 0.7632 & 0.6128 & 0.5752 & 0.7481 & 0.6090 & 0.6192 & 0.8346 & 0.6504 & 0.6248 & 0.8459 & 0.6429 & 0.6534 & 0.8571 & 0.6992 & \textbf{0.6726} & \textbf{0.8759} & \textbf{0.7105} \\
\hline
\end{tabular}
\label{tbl-building-eva}
\end{table*}

Based on the similarity matrix, one alternative option for mask clustering is spectral clustering~\cite{von2007tutorial}. We have evaluated it with different numbers of target clusters, noted as $K$, and compared our clustering method with it. For a fair comparison, we used two different values of $K$ for the spectral clustering method on each scene: 1) $K_1$: the building number of the annotated 3D scene, and 2) $K_2$: the number of global masks estimated by our clustering method. Note that $K_2$ is typically larger than $K_1$, as some of the global masks did not correspond to real 3D roofs. A more detailed explanation is given in Subsection \ref{sec-sub-buildingseg}. Specifically, $K_1$ has a value of $185/185, 119/119, 322/322, 266/266$, and $K_2$ has a value of $385/475, 202/376, 872/714, 646/641$ for Scene \#1, \#2, \#3, and \#4 with RGB/RGBH images, respectively.

The advantages of our clustering method over spectral clustering can also be concluded from Table~\ref{tbl-roof-eva} and Table~\ref{tbl-building-eva}. No matter on the 3D roof or building instance segmentation, using $K_1$ or $K_2$, with or without height information, our clustering method always reaches higher precision than spectral clustering. Visual comparison for Scene \#1 is shown in Fig.~\ref{fig-result-compare}. \cjz{From this comparison, we can observe that spectral clustering has the issue of under-segmentation when $K=K_1$. Since roof instance masks detected from the multi-view images are much more than the ground truth, and have prediction errors as well, thus the spectral clustering tends to incorrectly mix some roofs of attached buildings when $K=K_1$. Meanwhile, the spectral clustering has the issue of over-segmentation when $K=K_2$. Since $K_2$ is larger than the ground truth roofs in the 3D scene, it lost the ability to separate the correct and incorrect masks, leaving these masks separated.} In contrast, our instance mask clustering successfully segments roof instances precisely without specifying the number of target roofs. More visual comparison results can be found in the supplementary video.

It is worth pointing out that the last block named ``Ours with RGBH'' of Table~\ref{tbl-roof-eva} also shows that we achieve higher AP/AP50/AP75 on the 3D roof instance segmentation than on the aerial images (values are shown in Subsection~\ref{sec-sub-2d-roof-instance-segmentation}), because our occlusion-aware mask clustering suppresses false prediction from individual images and thus improves the overall segmentation precision.

\noindent\textbf{\cjz{Alternatives for image instance segmentation.}}
\cjz{The core contribution of our work is the multi-view framework with a new instance mask clustering, not the image instance segmentation. Besides Swin Transformer, our framework can incorporate any other image instance segmentation method as well. To demonstrate its compatibility, we have testified it with Mask R-CNN~\cite{he2017mask}, which is another widely-used image instance segmentation model. Table~\ref{tbl-multi-ortho-eva-mask-rcnn} shows a comparison between our framework based on multi-view images and the one based on orthophoto maps using Mask R-CNN. Similar to using Swin Transformer, the advantages of our multi-view method over the orthophoto-based method can also be found in this table. This demonstrates the effectiveness of our multi-view framework regardless of the chosen image instance segmentation method. Comparisons with the spectral clustering method and different MRF segmentation constraints using Mask R-CNN are shown in the supplementary document. For all these results, our method consistently achieves the highest accuracies. It is worth noting that most of the evaluation results for Mask R-CNN are lower than those of the recently developed Swin Transformer. }

\begin{table}
\footnotesize
\renewcommand{\arraystretch}{1.3}
\centering
\caption{\cjz{Comparison between our framework based on multi-view images and the one based on orthophoto maps using Swin Transformer.}}
\setlength{\tabcolsep}{1mm}
\linespread{1.3}
\begin{tabular}
{c|c c c|c c c}
\hline
 \multirow{2}{*}{Scene} & \multicolumn{3}{c|}{Orthophoto-based} & \multicolumn{3}{c}{Ours (multi-view)} \\
& AP & AP50 & AP75 & AP & AP50 & AP75 \\
\hline
\#1 & 0.6389 & 0.8432 & 0.6541 & \textbf{0.7130} & \textbf{0.8919} & \textbf{0.7730} \\
\#2 & 0.5975 & 0.7906 & 0.6059 & \textbf{0.6655} & \textbf{0.8403} & \textbf{0.6807} \\
\#3 & 0.5935 & 0.9006 & 0.6429 & \textbf{0.6671} & \textbf{0.9286} & \textbf{0.7174} \\
\#4 & 0.6199 & 0.8195 & 0.6391 & \textbf{0.6726} & \textbf{0.8759} & \textbf{0.7105} \\
\hline
\end{tabular}
\label{tbl-multi-ortho-eva}
\end{table}

\begin{table}
\footnotesize
\renewcommand{\arraystretch}{1.3}
\centering
\caption{\cjz{Comparison between our framework based on multi-view images and the one based on orthophoto maps using Mask R-CNN.}}
\setlength{\tabcolsep}{1mm}
\linespread{1.3}
\begin{tabular}
{c|c c c|c c c}
\hline
 \multirow{2}{*}{Scene} & \multicolumn{3}{c|}{Orthophoto-based} & \multicolumn{3}{c}{Ours (multi-view)} \\
& AP & AP50 & AP75 & AP & AP50 & AP75 \\
\hline
\#1 & 0.6751 & \textbf{0.8595} & 0.7189 & \textbf{0.7270} & \textbf{0.8595} & \textbf{0.7730} \\
\#2 & 0.5969 & 0.7176 & 0.5847 & \textbf{0.6202} & \textbf{0.7227} & \textbf{0.6471} \\
\#3 & 0.5475 & 0.8199 & 0.5714 & \textbf{0.6270} & \textbf{0.8354} & \textbf{0.7019} \\
\#4 & 0.6079 & \textbf{0.7895} & 0.6278 & \textbf{0.6117} & \textbf{0.7895} & \textbf{0.6353} \\
\hline
\end{tabular}
\label{tbl-multi-ortho-eva-mask-rcnn}
\end{table}

\noindent\textbf{MRF-based building segmentation.}
As shown in Fig.~\ref{fig-seg-new}, the orientation and distance constraints are important to achieve accurate 3D building segmentation. We have also evaluated the segmentation precision and recall by omitting one of them. The results are reported in Table~\ref{tbl-MRF-eva}, from which we can conclude that the distance constraint plays a more important role than the orientation constraint, but the best segmentation accuracy can only be achieved if they are both employed.

\begin{table*}
\footnotesize
\renewcommand{\arraystretch}{1.3}
\centering
\caption{\cjz{Comparison of using different segmentation constraints on 3D \textit{building} instance segmentation using Swin Transformer.}}
\setlength{\tabcolsep}{1mm}
\linespread{1.3}
\begin{tabular}
{c|c c c|c c c|c c c}
\hline
 \multirow{2}{*}{Scene} & \multicolumn{3}{c|}{Without the orientation constraint} & \multicolumn{3}{c|}{Without the distance constraint} & \multicolumn{3}{c}{With both constraints}\\
& AP & \quad AP50 & AP75 & AP & \quad AP50 & AP75 & AP & \quad AP50 & \quad AP75 \\
\hline
\#1 & 0.6438 & \quad 0.8541 & 0.7189 & 0.6856 & \quad 0.8673 & 0.7537 & \textbf{0.7130} & \quad \textbf{0.8919} & \quad \textbf{0.7730} \\
\#2 & 0.5605 & \quad 0.8235 & 0.5966 & 0.6378 & \quad 0.8325 & 0.6722 & \textbf{0.6655} & \quad \textbf{0.8403} & \quad \textbf{0.6807} \\
\#3 & 0.4252 & \quad 0.8540 & 0.3882 & 0.6177 & \quad 0.9224 & 0.6770 & \textbf{0.6671} & \quad \textbf{0.9286} & \quad \textbf{0.7174} \\
\#4 & 0.5289 & \quad 0.8308 & 0.5301 & 0.6176 & \quad 0.8722 & 0.6466 & \textbf{0.6726} & \quad \textbf{0.8759} & \quad \textbf{0.7105} \\
\hline
\end{tabular}
\label{tbl-MRF-eva}
\end{table*}

\subsection{Effects of parameters}
\label{subsec-para-eval}
Our method involves a few parameters, among which $\beta$ in the mask clustering step is the only parameter that we leave tunable for users (while all other parameters are fixed). In this subsection, we discuss how this parameter affects mask clustering.

\begin{figure*}[!t]
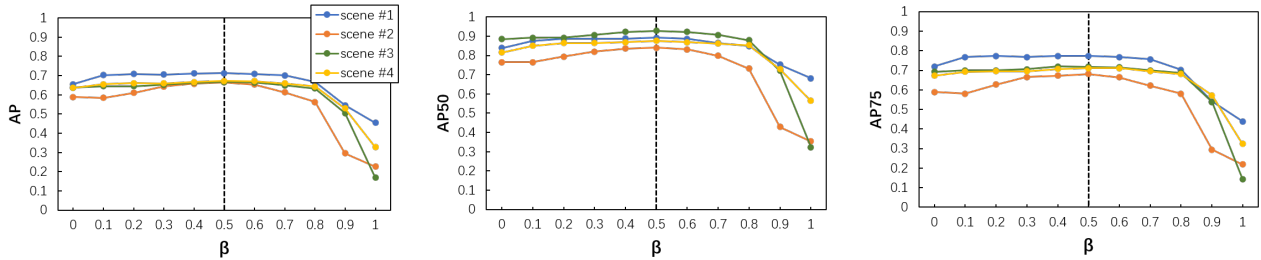

\centering
\begin{tabular}{@{}c@{\hspace*{0.01\linewidth}}c@{\hspace*{0.01\linewidth}}c@{\hspace*{0.01\linewidth}}c@{}}
	 \includegraphics[height=0.2\linewidth]{\imgfile{fig-beta-AP}} &
	 \includegraphics[height=0.2\linewidth]{\imgfile{fig-beta-AP50}} &
	 \includegraphics[height=0.2\linewidth]{\imgfile{fig-beta-AP75}}
\end{tabular}
\caption{\cjz{Evaluation of the segmentation results with different $\beta$ values using Swin Transformer. The AP (left), AP50 (middle), and AP75 (right) of the segmentation result are always close to the highest values when $\beta=0.5$.}}
	\label{fig-beta-comparison}
\end{figure*}

Intuitively, the meaning of $\beta$ parameter in our work is very similar to the threshold parameter of \emph{IoU} in many existing object detection works where a mask is considered to be correctly predicted when the \emph{IoU} between the detection mask and the ground truth is greater than this threshold. Empirically, this threshold is initially set to 0.5. Similarly, in our mask clustering, if the \emph{IoU} of the two local masks is greater than $\beta$, they should be considered to belong to the same group. That is, they represent the same roof instance. In this work, we initially set $\beta=0.5$ in all of our experiments. To determine the optimal value for $\beta$, we experimented with different values in all 4 scenes. As we can see from Fig.~\ref{fig-beta-comparison}, the AP, AP50, and AP75 are always close to the highest values when $\beta=0.5$. Note that slightly increasing/reducing $\beta$ reduces/increases the confidence values but barely affects the ordering of the masks. This reveals that our mask clustering is tolerant to the $\beta$ parameter.

\begin{figure*}
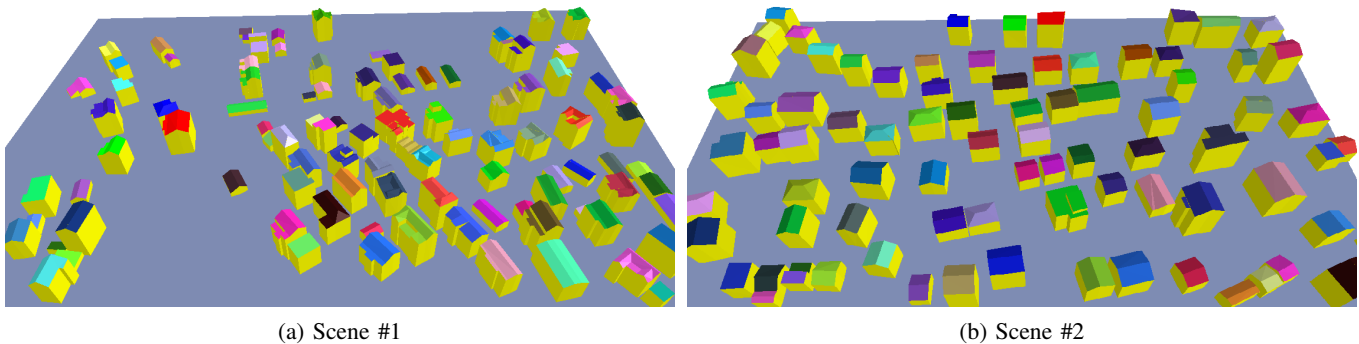

\centering
\begin{tabular}{@{}c@{\hspace*{0.01\linewidth}}c@{}}
\includegraphics[width=0.49\linewidth]{\imgfile{fig-simple4}} &
\includegraphics[width=0.49\linewidth]{\imgfile{fig-simple5}} \\
{\small{(a) Scene \#1}} &
{\small{(b) Scene \#2}}
\end{tabular}
\caption{\cjz{Automatic simplification of buildings in two large urban scenes. The results are obtained by applying RANSAC for plane extraction followed by plane regularization.}}
\label{fig-application}
\end{figure*}

\subsection{Running time}
The training took around 83 hours with 2000 epochs. The overall running time for segmenting a scene varied from 6 to 8 minutes, depending on the scene size,  image resolution, and the number of images. The MRF optimization takes around 2.5 minutes on average for each scene. \cjz{The computational complexity of multi-view image generation (only for scenes without drone images), heightmap generation, 3D vertex projection for the overlapping computation, and back projection for the clustered instance masks are $O(N*k)$, where $N$ is the number of faces in this 3D urban model, and $k$ is the number of multi-view images. $N$ could be a large number, but these computations are fully accelerated using the GPU, only taking less than 1 minute for each urban scene. The instance mask clustering has a computational complexity of $O(n)$, where $n$ is the number of instance masks, thus it takes less than 1 second in total for each urban scene.} It is much faster than many other traditional clustering methods.

\subsection{Discussions}
\noindent\textbf{Applications.} We have implemented a simple building simplification prototype using a RANSAC-based plane fitting~\cite{Schnabel2007} and plane regularization. Based on our 3D building instance segmentation, 3D buildings in large urban scenes are automatically simplified, as shown in Fig.~\ref{fig-application}. In such an application, instance segmentation is obligatory, as semantic segmentation is insufficient to separate individual buildings. We believe our 3D building instance segmentation method can benefit more smart city applications, such as urban planning.

\begin{figure}
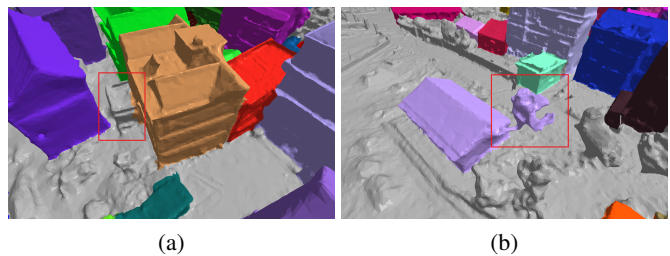

\centering
\begin{tabular}{@{}c@{\hspace*{0.01\linewidth}}c@{}}
\includegraphics[width=0.49\linewidth]{\imgfile{fig-error4}} &
\includegraphics[width=0.49\linewidth]{\imgfile{fig-error3}} \\
{\small{(a)}} &
{\small{(b)}}
\end{tabular}
\caption{\cjz{Two failure cases of our method. (a) A gated shelter was not segmented as part of the roof in the image segmentation stage. (b) A tall tree was segmented as a roof instance because it has a similar height and texture as its nearby building. }}
\label{fig-failure}
\end{figure}

\noindent\textbf{Limitations.}
Since our approach falls into the multi-view paradigm, it relies on the quality of the 2D instance segmentation. Roof types that do not exist in the training dataset may not be precisely segmented. Two such examples are shown in Fig.~\ref{fig-failure}. In Fig.~\ref{fig-failure} (a), a gated shelter was not reliably detected, and in Fig.~\ref{fig-failure} (b), a tall tree was segmented as a part of the nearby building. Enriching the training dataset may partially solve this issue. In addition, developing a neural network dedicated to separating roofs and trees may produce more reliable instance segmentation results.

\label{sec:results}

\section{Conclusion and Future Work}
We have presented a multi-view framework for instance segmentation of 3D urban buildings. Based on occlusion-aware similarity matrices, a novel instance mask clustering method is proposed to eliminate the mask ambiguities among multi-view images. To further improve segmentation accuracy, roofs (instead of buildings) are firstly segmented, and RGB images are enriched with heightmaps. Our method takes full advantage of the multi-view framework to precisely segment 3D buildings in large urban scenes.

We have collected and annotated an RGBH drone imagery dataset and a 3D building instance segmentation dataset, named \datasetname. We believe the new dataset could benefit research in 3D instance segmentation for various urban applications. Since most of the state-of-the-art learning-based 3D instance segmentation methods focus on indoor scenes, our multi-view instance segmentation framework explores a new avenue for large outdoor scenes.

\noindent\textbf{Future directions}. \cjz{Our work focuses on buildings because they are the most important ingredients in the urban environment. One future direction is to extend our multi-view 3D instance segmentation framework to other urban objects and even indoor scenes. Drone images and the reconstructed 3D models may suffer from various degradation (such as noises), it is worth investigating the robustness of our method in such degraded scenarios\cite{hong2018augmented}. Finally, applying our method to 3D point clouds of urban scenes could be an interesting future direction as well.}



\label{sec:conclusion}

\section*{Acknowledgment}
The authors would like to thank Quzhou Southeast Flysee Technology Ltd. for providing drone images, 3D urban models and parts of 2D annotations. This work was supported in part by National Natural Science Foundation of China (62172367), and in part by Natural Science Foundation of Zhejiang Province (LGF22F020022).

\ifCLASSOPTIONcaptionsoff
  \newpage
\fi



\bibliographystyle{IEEEtran}
\bibliography{IEEEabrv,./InstanceSegmentation}
\end{document}